\documentclass[lettersize,journal]{IEEEtran}
\usepackage{amsmath,amsfonts}
\usepackage{algorithmic}
\usepackage{algorithm}
\usepackage{array}
\usepackage[caption=false]{subfig}
\usepackage{textcomp}
\usepackage{stfloats}
\usepackage{url}
\usepackage{verbatim}
\usepackage{graphicx}
\usepackage{cite}

\usepackage[table,xcdraw]{xcolor}
\definecolor{darkgreen}{RGB}{0, 180, 0}
\definecolor{darkred}{RGB}{180, 0, 0}
\definecolor{pinktab}{RGB}{255, 240, 240}
\definecolor{ablatab}{gray}{.9}
\usepackage{multirow}
\usepackage{booktabs}
\usepackage[hidelinks]{hyperref}
\usepackage{eso-pic}

\begin{document}
\newcommand{\ie}{\textit{i}.\textit{e}.}
\newcommand{\eg}{\textit{e}.\textit{g}.}
\newcommand{\etal}{\textit{et~al}.}
\newcommand{\NA}{---}
\bstctlcite{IEEEexample:BSTcontrol}

\title{Conditioning Residuals for Diffusion Models \\ via Representation Feedback}

\author{Weilai~Xiang,
        Hongyu~Yang,~\IEEEmembership{Member,~IEEE,}
        Di~Huang,~\IEEEmembership{Senior~Member,~IEEE,}
        and~Yunhong~Wang,~\IEEEmembership{Fellow,~IEEE}% <-this % stops a space
\thanks{Corresponding authors: Hongyu Yang (e-mail: hongyuyang@buaa.edu.cn) and Yunhong Wang (e-mail: yhwang@buaa.edu.cn).}%
\thanks{W. Xiang and Y. Wang are with the State Key Laboratory of Virtual Reality Technology and Systems, School of Computer Science and Engineering, Beihang University, Beijing 100191, China.}%
\thanks{H. Yang is with the Institute of Artificial Intelligence, Beihang University, Beijing 100191, China.}%
\thanks{D. Huang is with the School of Computer Science and Engineering, Beihang University, Beijing 100191, China.}%
}

\markboth{IEEE Transactions on Multimedia}%
{Xiang \MakeLowercase{\textit{et al.}}: Conditioning Residuals for Diffusion Models via Representation Feedback}


\AddToShipoutPictureFG*{
  \AtPageLowerLeft{
    \put(0,16){
      \makebox[\paperwidth][c]{
        \parbox{\textwidth}{
          \centering
          \scriptsize This work has been submitted to the IEEE for possible publication. \\
          Copyright may be transferred without notice, after which this version may no longer be accessible.
        }
      }
    }
  }
}

\maketitle

\begin{abstract}
Diffusion models now serve as a common foundation for multimedia generation, and useful intermediate representations emerge during their generative training. Standard architectures, however, propagate these representations through the main feature stream, without explicitly reintroducing their encoded semantics to later denoising layers. Meanwhile, such backbones already provide a conditioning pathway for global modulation by predefined inputs. This work examines whether this native pathway can also route internally inferred semantics as evolving, sample-dependent cues.
We propose Conditioning Residuals, a lightweight feedback mechanism that converts aggregated features into residuals added to condition embeddings. By feeding back compact feature summaries, it provides adaptive generative guidance and encourages a tighter semantic bottleneck, without external encoders, auxiliary objectives, or sampling-time changes. It supports feedback at one or multiple depths in UNet and DiT backbones, with negligible overhead.
Across diffusion formulations, backbone configurations, and datasets, experiments show consistent gains in generative performance, along with stronger representations in downstream linear probing and segmentation. Mechanistic analyses reveal improved generative training dynamics and reshaped feature structure, suggesting a grounded, generalizable way to enhance diffusion backbones from within.
\end{abstract}


\section{Introduction}
\IEEEPARstart{D}{iffusion} models \cite{ddpm, edm, sde} and their flow-based variants \cite{flow, rf} (collectively referred to as \textit{diffusion models}) have become central to modern generative AI, particularly for high-fidelity visual synthesis \cite{adm, ldm}. Their impact has extended to diverse multimedia domains, powering image/video generation and editing \cite{kontext, wan, waver}, unified multimodal models \cite{d-dit, metaqueries, tuna-2}, world and action models \cite{pi0, lingbot, dreamzero}, and generative compression and transmission \cite{OneDC, GNVC-VD, GVC}. This breadth positions diffusion models as a \textit{shared foundation} for content representation, synthesis, and communication.

Despite this ubiquity, the underlying diffusion models still face persistent challenges, such as slow convergence \cite{repa} and imperfect visual details \cite{hallu}. Many recent efforts have turned to system-level additions: richer input conditions \cite{dalle3}, external semantic guidance \cite{rcg}, or task-specific modules \cite{hand}. While effective, such strategies often operate on the surrounding interfaces rather than the generator's internal mechanisms. This motivates a complementary direction: revisiting the shared diffusion backbone and improving it \textit{from within}.

A natural entry point is the model's intermediate representations. Prior studies reveal that discriminative features emerge from generative pre-training, capturing both dense \cite{ddpm-seg} and high-level \cite{ddae} semantics. Recent work further shows that enhancing them via alignment with external encoders \cite{repa} or self-distillation \cite{sra} can improve training efficiency and generation quality. These findings suggest that such representations are not merely by-products, but a key interface for observing and intervening in a model's denoising dynamics \cite{navigating}.

Yet these representations may remain structurally underutilized. In typical architectures (\eg, UNet \cite{unet} and DiT~\cite{dit}), they are treated as standard feature maps or patch tokens that are sequentially propagated through layers, without explicit semantic routing. Once formed at certain depths, their semantic content is not selectively preserved for later consumption, but its influence depends mainly on layer-wise propagation and residual mixing \cite{attnres}. The question is thus not only whether useful representations exist, but whether the architecture enables them to contribute more effectively to generation.

% ==================== Figure 1: Conceptual Overview ====================
\begin{figure}[t!]
\centering
\includegraphics[width=\linewidth]{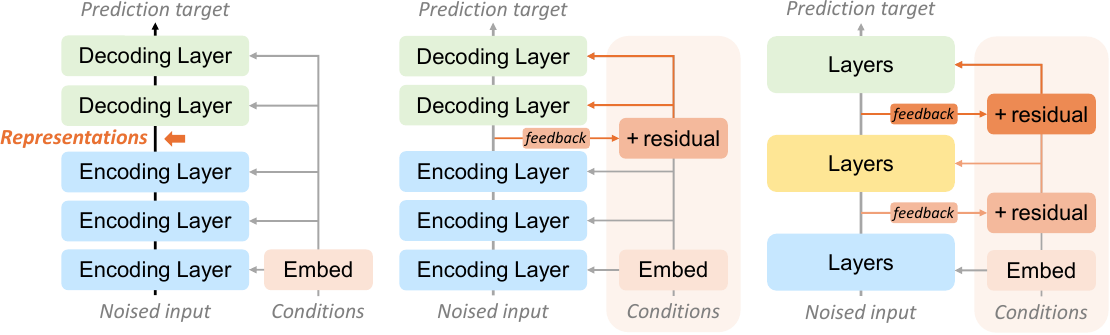}
\caption{
    \textbf{Conceptual Overview of Conditioning Residuals.}
    \textit{Left:} Standard backbones contain useful representations, while the conditioning pathway is driven only by predefined inputs.
    \textit{Middle:} Single-shot feedback converts one representation into a conditioning residual for later layers.
    \textit{Right:} Multi-shot feedback repeats this across depths, forming a dynamic conditioning stream.
}
\label{fig:intro}
\end{figure}
% ==================== Figure 1: Conceptual Overview ====================

Beyond the main representation stream, diffusion architectures also contain a dedicated conditioning pathway that injects timestep, class, or other conditions into layers, often through adaptive normalization \cite{adm, dit}. This pathway distinguishes them from conventional visual encoders \cite{vit}: it provides a native interface for global modulation, making the generation process governed by control signals. Similar pathways also appear in broader diffusion-based multimedia systems \cite{sd3, PrimePSegter, DivDiff}.
Therefore, beyond routing features more effectively within the main stream \cite{dar}, we identify a \textbf{feedback} route: intermediate representations, already carrying semantically rich, decodable information, can serve as dynamic conditions analogous to class embeddings, yet derived from the model itself.

Motivated by this, we propose Conditioning Residuals, a mechanism for self-guidance within the forward pass, without requiring major framework overhauls or external knowledge. As illustrated in Fig.~\ref{fig:intro}, we aggregate and reroute intermediate features back into the existing conditioning pathway, allowing the model to leverage internally encoded semantics to guide its own decoding layers. These residuals do not replace the original conditions; instead, they augment them and make the resulting control signal progressively sample-aware.

This design brings three coupled benefits.
\textbf{(1) Information Flow.} Acting as a side branch, Conditioning Residuals shorten the path from where information is formed to where it is consumed, reducing reliance on sequential propagation.
\textbf{(2) Generation.} By connecting hidden states back to conditioning, they establish a feedback loop in which generation is shaped by its own evolving states. As generative models benefit from informative conditions \cite{scdm}, these self-contained semantic cues provide additional guidance and may help translate improved representations into improved generation.
\textbf{(3) Representation.} Since decoding layers explicitly depend on compact feature summaries, the model is encouraged to concentrate high-level information for reuse. This can reshape the layer-wise semantic hierarchy toward a tighter bottleneck \cite{soda}, making semantics less broadly distributed across layers \cite{ddpm-seg}.

We implement this mechanism as a lightweight plug-in for architectures with explicit conditioning pathways, using UNet and DiT as representative backbones. Concretely, intermediate features at feedback depths are globally pooled, projected with time-dependent modulation, and added to condition embeddings, which are then broadcast to subsequent layers. We study two variants: \textbf{single-shot} feedback at a designated depth as the main configuration, and \textbf{multi-shot} feedback as a generation-oriented, complementary extension with repeated updates.

We isolate Conditioning Residuals as the sole architectural change under matched training and evaluation protocols, and evaluate it across \textbf{17 controlled settings} spanning pixel-space CIFAR-10/100 generation with varied model formulations and UNet scales, as well as latent-space DiTs on ImageNet. Across these settings, single-shot feedback consistently improves FID and achieves linear probing gains in nearly all cases, with negligible overhead. We further examine semantic segmentation on ADE20K and VOC2012, and conduct mechanistic analyses to understand how feedback reorganizes learned states across depth and training. Multi-shot feedback also improves over baselines and often outperforms single-shot results.

Our main contributions are threefold. (1) We identify an underexplored opportunity in diffusion architectures and formulate representation-as-condition feedback as an internal mechanism for improving information flow. (2) We instantiate this idea as Conditioning Residuals for UNet and DiT, covering both single-shot and multi-shot feedback. (3) We validate the mechanism across models, backbones, and datasets, showing dual gains in generation and representation quality, as well as additional generative benefits from repeated feedback.

\section{Related Work}
\subsection{Representations in Diffusion Models}
Diffusion models can be repurposed for discriminative tasks through intermediate representations, including dense prediction (\eg, semantic segmentation \cite{ddpm-seg, VPD}) and image recognition \cite{ddae}. DDAE further links generative quality to feature quality, framing them as unified learners for content generation and understanding \cite{ddae}. From a denoising-autoencoding view \cite{mae, ldae}, these models can be interpreted as a composition of an \textit{encoder} that recovers information from corrupted inputs and a \textit{decoder} that predicts the target from it. This view helps explain why rich, transferable semantics emerge.

This link also makes representation quality a complementary measure of a diffusion model's quality. REPA relates the alignment between hidden states and strong visual encoders to generation performance \cite{repa}, while iREPA further emphasizes spatial structure as another indicator \cite{irepa}. These studies suggest that high-level semantics and dense information are valuable aspects for evaluating diffusion models.

Recent studies thus target these representations or their flow to improve diffusion models. For the former, existing methods strengthen hidden states through external alignment \cite{repa}, self-distillation \cite{sra}, and hybrid generative-discriminative designs \cite{navigating}. DAR \cite{dar} instead revisits how information is propagated across layers, adapting AttnRes-style residual connections \cite{attnres} to diffusion models. Both categories operate primarily within the feature stream: the former enhances hidden state quality, whereas the latter changes cross-layer accumulation. Our method is not another modification confined to the feature stream; instead, it adds a \textit{side branch} that extracts intermediate representations from the stream and repurposes them as conditions to explicitly modulate later denoising layers.

\subsection{Conditioning for Diffusion Models}
Diffusion architectures provide standard pathways for global conditional information. Inputs such as timestep, class label, or pooled text features can be encoded and combined into a compact signal. Modern models inject it into denoising blocks through adaptive normalization, such as AdaGN in ADM-style UNet \cite{adm} and AdaLN in DiT \cite{dit}, where the shared signal is mapped by each block to its own modulation parameters.

Such information is known to improve generation quality, and conditional models commonly outperform unconditional ones \cite{scdm}. Recent methods enrich the input with extra semantic cues: RCG \cite{rcg} uses representations from pre-trained self-supervised encoders, while SCDM \cite{scdm} and SGDM \cite{sgdm} derive pseudo-labels by clustering such representations. However, this route assumes a compatible encoder that can produce high-quality features for the target domain. It can also be restrictive when representations matched to the desired output are not readily available at inference (\eg, in image editing \cite{nullinv, instructpix2pix}).

This motivates a self-contained alternative that \textit{internalizes} semantic conditioning: the model derives sample-dependent cues from its own evolving intermediate states. Our method follows this direction by constructing feedback residuals added to the conditioning pathway, forming a dynamic conditioning stream without additional inputs or auxiliary regularization.

\section{Preliminaries}
\subsection{Diffusion Formulations}
We use \textit{diffusion models} to encompass both diffusion/score-based models \cite{ddpm, edm, sde} and their closely related flow-based variants \cite{flow, rf}. These models construct paths that progressively transform data into noise via a forward process over $t\in[0,T]$:
\begin{equation}
    x_t = \alpha_t x_0 + \sigma_t \epsilon \quad \text{where} ~ \epsilon \sim \mathcal{N}(0, I), \label{eq:forward}
\end{equation}
with schedules for $\alpha_t, \sigma_t$ such that $x_0$ follows $p_{\rm data}$ and $x_T$ approaches a Gaussian prior. To reverse this, an ordinary differential equation (ODE) can be formulated \cite{sde, flow}:
\begin{equation}
    \mathrm{d} x_t = v(x_t, t) \mathrm{d} t,
\end{equation}
where the induced distribution matches the marginal $p_t(x_t)$ derived from Eq.~(\ref{eq:forward}). The velocity field $v$ can be approximated by a time-conditioned network $v_\theta$, which, once trained, enables ODE solvers to sample data from noise.

Learning such reverse dynamics is closely related to denoising autoencoding and score matching \cite{sde}, allowing the training objective to be reparameterized into flexible forms. Different formulations of diffusion models may vary in prediction target, noise schedule, and sampler. In this paper, we examine three representative ones, providing a brief overview of each.

\textbf{DDPM} \cite{ddpm} operates over discrete timesteps with a variance-preserving schedule, \ie, $x_t = \alpha_t x_0 + \sqrt{1 - \alpha_t^2} \epsilon$, where $\alpha_t = \sqrt{\smash{\prod\nolimits_{i=1}^{t}} (1 - \beta_i)}$ for a predefined linear $\beta$ schedule. The network is trained using the $\epsilon$-prediction objective $\lVert \epsilon_\theta(x_t, t) - \epsilon \rVert_2^2$. We use DDIM-style deterministic sampling \cite{ddim}.

\textbf{EDM} \cite{edm} is a continuous-time model based on the variance-exploding parameterization, \ie, $x_t = x_0 + \sigma_t \epsilon$, where $\sigma_t \in [0.002, 80]$. The model predicts the original image and is trained with a denoising objective based on $\lVert D_\theta(x_t, t) - x_0 \rVert_2^2$. The second-order Heun solver is used for efficient sampling.

\textbf{FM} (Flow Matching) \cite{flow} uses a linear interpolation between data and noise, \ie, $x_t = (1-t) x_0 + t \epsilon$, with $t \in [0,1]$. The network is trained using a $v$-prediction loss $\lVert v_\theta(x_t, t) - (\epsilon - x_0) \rVert_2^2$. The RK45 ODE solver is used for pixel-space sampling \cite{rf}, while Euler's method is used for the latent space \cite{lightning}.

\subsection{Backbone Architectures}
To support comprehensive evaluation, we separate network backbones from their model formulations. This lets us flexibly combine them, yielding a broader testbed to examine whether the plug-in is \textit{formulation- and backbone-agnostic}.

\textbf{UNet backbones} date back to early diffusion models, where the timestep is encoded by sinusoidal embeddings and injected into residual blocks after an MLP. In our experiments, we use two common configurations: the original DDPM UNet by Ho \etal~\cite{ddpm}, denoted as \texttt{UNet\,(Ho)}, and the DDPM++ version from Song \etal~\cite{sde}, denoted as \texttt{UNet\,(Song)}. The latter increases model depth and employs improved residual blocks. These variants mainly inject timestep embeddings via additive conditioning, a route later refined into ADM-style AdaGN \cite{adm}, modulating features by scale and shift parameters.

\textbf{DiT backbones} inherit the token-based design of Vision Transformers \cite{vit}. Additionally, timestep and other conditions are injected through AdaLN \cite{dit}, analogous to UNet's AdaGN. Such backbones have been effective for latent-space generation \cite{ldm}, including with FM formulations \cite{sit}. Among them, we choose LightningDiT as the representative, since its off-the-shelf VA-VAE provides an improved latent space that enables faster DiT convergence \cite{lightning}, while the DiT itself still follows its established training recipe without representation-level intervention. Thus, we can study our method as the sole modification on top of an established modern baseline.

\textbf{Unified view.} Despite architectural differences, UNet- and DiT-style backbones cover the dominant designs underlying modern diffusion models \cite{wan, d-dit}, and share similar conditioning pathways that are reused across layers. Our mechanism relies on this common interface rather than a specific block design, making it applicable to these backbones and conceptually straightforward to extend to other related variants \cite{sd3}.

\section{Methodology}
\noindent
To present Conditioning Residuals, we first define the representations, conditioning pathway, and residual augmentation, then describe single-/multi-shot feedback, and explain how feedback depths are chosen and why the cost is negligible.

\subsection{Overview}
Let the diffusion network $v_\theta$ (or $\epsilon_\theta, D_\theta$ under other parameterizations) be a sequence of layers that propagate activations while receiving a global condition embedding. Let $\mathbf{h}_l$ denote the activation after the $l$-th layer, where $\mathbf{h}_l \in \mathbb{R}^{C_l \times H_l \times W_l}$ for UNets and $\mathbf{h}_l \in \mathbb{R}^{N \times D}$ for DiTs.
For UNets, we absorb skip connections into ordinary feature propagation, since they pass feature maps through non-selective links and remain within the feature stream rather than entering the conditioning pathway.
Let $\mathbf{e}_t=\phi(t) \in \mathbb{R}^{d}$ be the timestep embedding, where $\phi$ typically contains a fixed encoding followed by a learned two-layer MLP. An optional condition $y$, such as a class label, is encoded separately and combined with $\mathbf{e}_t$ when present. We denote the resulting global condition embedding by $\mathbf{e} \in \mathbb{R}^{d}$.

A standard backbone can then be written abstractly as
\begin{equation}
\mathbf{h}_l = B_l(\mathbf{h}_{l-1}; \mathbf{e}), \quad l=1,\ldots,L,
\end{equation}
where each $B_l$ consumes $\mathbf{e}$ through additive conditioning, AdaGN, or AdaLN. Although different layers map $\mathbf{e}$ to layer-specific modulation parameters, the embedding itself is fixed by the input conditions before these activations are computed.

Conditioning Residuals augment this interface by adding feedback-derived terms to the condition embedding. When the forward pass reaches a feedback depth $m$, the output activation $\mathbf{h}_m$ is first aggregated into a compact summary and then transformed into an additive residual:
\begin{equation}
\mathbf{s}_m = A_m(\mathbf{h}_m), \quad
\Delta \mathbf{e}_m = F_m(\mathbf{s}_m,t).
\end{equation}
Here, $A_m$ denotes a lightweight aggregation operator over $H_m \times W_m$ spatial locations or $N$ patch tokens, and $F_m$ denotes a feedback head that maps the summary $\mathbf{s}_m$, conditioned on the current timestep $t$, to the same $d$-dimensional space as $\mathbf{e}$.

The residuals are generated online during the forward pass, and $\Delta \mathbf{e}_m$ is immediately propagated to subsequent layers for consumption, from layer $m+1$ onward. For a set of selected feedback depths $\mathcal{M}$, the resulting condition embedding used by the $l$-th layer is therefore
\begin{equation}
\tilde{\mathbf{e}}_l
= \mathbf{e} +
\sum_{\substack{m \in \mathcal{M} \\ m<l}}
\Delta \mathbf{e}_m,
\quad
\mathbf{h}_l = B_l(\mathbf{h}_{l-1}; \tilde{\mathbf{e}}_l).
\end{equation}

Algorithm~\ref{algorithm} summarizes the generalized forward-pass update, where single-shot feedback corresponds to a singleton $\mathcal{M}$, and multi-shot feedback uses multiple depths and progressively enriches the embedding. This augmentation leaves training, sampling, and native layer operations unchanged. The only additions are $A_m$ and $F_m$. We instantiate these components for single-shot and multi-shot feedback as follows.

% ==================== Figure 2: Implementation ====================
\begin{figure}[t!]
\centering
\includegraphics[width=0.85\linewidth]{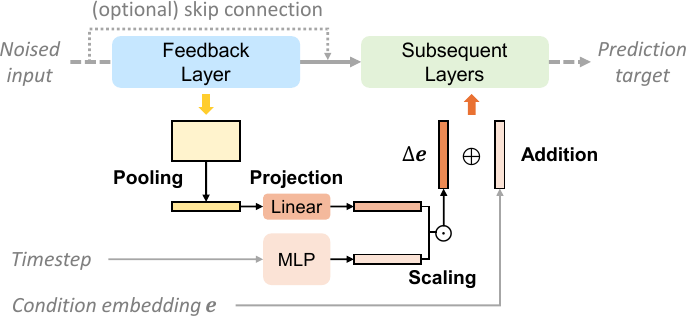}
\caption{
    \textbf{Illustration of Representation Feedback.}
    At a feedback depth,~the activation is pooled into a summary vector, projected into the embedding space by a linear layer, and modulated by the timestep to form a residual.
}
\label{fig:implementation}
\end{figure}
% ==================== Figure 2: Implementation ====================

\subsection{Single-shot Conditioning Residuals}
Single-shot feedback is the minimal form of our approach. Given one designated feedback depth $m$, the backbone remains unchanged up to layer $m$. After $B_m$ produces the selected representation $\mathbf{h}_m$, a feedback branch aggregates and condenses its semantic content into a compact summary, which is then transformed into a conditioning residual consumed by layers $m+1,\ldots,L$. In this subsection, we focus on the feedback operation itself, while the choice of $m$ is discussed later.

\textbf{Aggregation.} We use parameter-free global average pooling. For UNets, spatial averaging gives $\mathbf{s}_m \in \mathbb{R}^{C_m}$; for DiTs, token averaging gives $\mathbf{s}_m \in \mathbb{R}^{D}$. The resulting summary vector acts as a bottleneck, encouraging reusable high-level information to concentrate. Although more expressive choices such as attention modules may also work, we keep it simple to focus purely on whether the aggregated information is useful.

\textbf{Feedback.} The aggregated summary is then transformed through a lightweight feedback head, as shown in Fig.~\ref{fig:implementation}. This head first projects the summary to the embedding space, and then modulates it with a time-dependent scaling vector:
\begin{equation}
\Delta \mathbf{e}_m = P_m(\mathbf{s}_m) \odot \phi_m(t),
\label{eq:shot}
\end{equation}
where $P_m$ is a $\mathbb{R}^{C_m}\rightarrow\mathbb{R}^{d}$ projection for UNets, or $\mathbb{R}^{D}\rightarrow\mathbb{R}^{d}$ for DiTs; $\phi_m(t)\in\mathbb{R}^{d}$ is predicted by a separate timestep embedder, and $\odot$ denotes element-wise multiplication. Following the same minimal design, we implement $P_m$ as a single linear layer, while other choices are possible. We zero-initialize $P_m$ in the spirit of \cite{dit}, so that $\Delta \mathbf{e}_m=\mathbf{0}$ at initialization and the network exactly recovers the baseline behavior.

We use the time-dependent scaling because direct addition can be suboptimal for generation. Different timesteps may require varying amounts of semantic guidance, and the aggregated feature itself may also be less reliable outside the range where it is most informative. The scaling vector $\phi_m(t)$ therefore allows the model to adjust the strength and channel-wise pattern of the feedback according to the current timestep.

As shown in Fig.~\ref{fig:implementation}, our single-shot implementation adds only one linear layer and one timestep embedder. No auxiliary objective, extra forward pass, external representation pipeline, or sampling-time modification is introduced. Both the training recipe and sampling procedure remain identical to the baseline.

% ==================== Algorithm 1: Pseudocode ====================
\begin{algorithm}[t!]
\caption{Forward Pass with Conditioning Residuals}
\label{algorithm}
\small
\noindent\textbf{Input:} $x_t$, $t$, $y$, $\{B_l\}_{l=1}^{L}$, $\mathcal{M}$,
aggregation operators $\{A_m\}_{m\in\mathcal{M}}$, and feedback heads
$\{P_m,\phi_m\}_{m\in\mathcal{M}}$
\begin{algorithmic}[1]
\STATE $\mathbf{h}_0 \gets \mathrm{InputEmbed}(x_t)$
\STATE $\mathbf{e} \gets \mathrm{CondEmbed}(t,y)$
\FOR{$l=1,\ldots,L$}
    \STATE $\mathbf{h}_l \gets
    B_l(\mathbf{h}_{l-1}; \mathbf{e})$
    \IF{$l \in \mathcal{M}$}
        \STATE $\mathbf{s}_l \gets A_l(\mathbf{h}_l)$
        \STATE $\Delta\mathbf{e}_l \gets
        P_l(\mathbf{s}_l)\odot\phi_l(t)$
        \STATE $\mathbf{e} \gets \mathbf{e}+\Delta\mathbf{e}_l$
    \ENDIF
\ENDFOR
\STATE \textbf{return} $\mathrm{OutputHead}(\mathbf{h}_L)$
\end{algorithmic}
\end{algorithm}
% ==================== Algorithm 1: Pseudocode ====================

\subsection{Multi-shot Conditioning Residuals}
Multi-shot feedback extends the single-shot design from one designated depth to a small set of depths $\mathcal{M}$. Since the most effective feedback position may vary slightly across settings, this extension relaxes the reliance on a single choice and allows representations of varying semantic granularity formed at different layers to provide complementary updates.

Concretely, during the forward pass, whenever the network reaches a depth $m \in \mathcal{M}$, its activation $\mathbf{h}_m$ is aggregated and transformed into $\Delta\mathbf{e}_m$. The running condition embedding is then updated immediately, so that the layers after $m$ consume the augmented embedding, and later feedback at $m'>m$ is based on activations already shaped by earlier updates.

Instead of caching multi-layer features and fusing them into a single residual, this design scales the single-shot method through repeated, stage-wise conditioning updates, and thus does not require a separate feature fusion module.

For residual construction, UNets simply repeat the feedback head in Eq.~(\ref{eq:shot}) at each selected depth. For DiTs, whose blocks share the same dimension $D$, we use a shared projection across selected depths but separate time modulation:
\begin{equation}
\Delta\mathbf{e}_m = P(\mathbf{s}_m)\odot\phi_m(t),
\quad
\Delta\mathbf{e}_{m'} = P(\mathbf{s}_{m'})\odot\phi_{m'}(t).
\label{eq:shot_shared}
\end{equation}
Here, the shared $P$ maps summaries from different DiT layers into a common residual space, while each keeps its own $\phi_m$, allowing the strength to remain depth- and timestep-dependent.

Since multi-shot feedback distributes the bottleneck across depths, we do not focus on representation quality at an individual layer. We therefore evaluate it mainly as a generation-oriented extension, testing whether repeated conditioning updates can further improve generative performance.

\subsection{Feedback Placement and Cost}
Feedback placement needs to balance semantic encoding and decoding consumption. Very early layers may lack stable high-level information, whereas very late layers leave little room for influence, motivating us to place feedback in an intermediate semantic region. The best linear probing layer in the corresponding baseline can provide a useful reference, but it does not always coincide with the best feedback layer.

Rather than performing an exhaustive full-training search over all layers, we restrict depth selection to a small set of candidates and use \textit{early convergence} as a practical diagnostic. Specifically, we conduct short profiling runs and select the candidate with the lowest early training loss. This is motivated by the observation that an effective feedback position should improve training dynamics, and can therefore be identified before full convergence. The selected depths are then fixed for full comparisons and summarized by backbone in Table~\ref{tab:placement}.

% ==================== Table 1: Config + Cost ====================
\begin{table}[t]
    \centering
    \caption{Feedback Placement and Overhead}
    \label{tab:placement}
\resizebox{\linewidth}{!}{
\begin{tabular}{lcccc}
\hline
Backbone    & Total $L$ & Single-shot $m$ & Multi-shot $\mathcal{M}$ & Single-shot Param. \\
\hline
UNet (Ho)   & 12 &  6--7  & \{4, 8\}   & +0.46M (1.3\%) \\
UNet (Song) & 15 &  8--9  & \NA        & +0.46M (0.8\%) \\
DiT-B       & 12 &   9    & \{4, 8\}   & +1.38M (1.1\%) \\
DiT-L       & 24 & 17\,(u), 18\,(c) & \{14, 18\} & +2.36M (0.5\%) \\
DiT-XL      & 28 &   20   & \NA        & +2.95M (0.4\%) \\
\hline
\end{tabular}
}

\vspace{1.5ex}
\begin{minipage}{0.95\linewidth}
Since the best $m$ may vary slightly across settings, the ranges summarize these neighboring choices. Exact CIFAR choices appear in supplementary Tab.~\ref{tab:cifar_epochs}. (u)/(c) denote unconditional/conditional ImageNet settings. A dash indicates that multi-shot feedback was not evaluated for that backbone.
\end{minipage}
\end{table}
% ==================== Table 1: Config + Cost ====================

For multi-shot feedback, we keep $\mathcal{M}$ fixed per backbone. We prioritize representative backbones across scales with sufficient room for improvement. On 12-layer backbones, including \texttt{UNet\,(Ho)} and DiT-B, we use $\mathcal{M}=\{4,8\}$, which gives two roughly evenly spaced feedback points. The four-layer offset keeps the two points within related high-level stages, while also separating them enough to reduce redundancy. For DiT-L, we use double-shot $\mathcal{M}=\{14,18\}$, where the two points span the main part of the semantic region. This fixed schedule avoids setting-specific tuning and directly examines whether repeated updates can provide further gains.

The added cost remains small. Single-shot feedback adds a linear projection and a two-layer timestep embedder, corresponding to three more linear layers. Double-shot replicates this lightweight head and optionally reuses the projection, adding five or six linear layers. Wall-clock cost is effectively unchanged: the mean training overhead remains below 0.6\%; single-shot adds at most 0.5\% sampling overhead, and double-shot remains around 1\%; see the supplementary material.

\section{Experiments}
\noindent
We evaluate our approaches from two aspects: empirical effects and underlying mechanisms. We first test whether single-shot feedback concurrently improves generation and representation quality across 17 settings, and then examine whether the multi-shot extension brings further gains on 8 settings. Finally, we investigate how the feedback reshapes layer-wise feature hierarchy and how it affects training dynamics.

\subsection{Experimental Setup}
\textbf{Settings.} We evaluate the single-shot variant on 17 controlled settings. In pixel space, we use CIFAR-10/100 with three formulations (DDPM, EDM, and FM) and two backbones (\texttt{UNet\,(Ho)} and \texttt{UNet\,(Song)}), resulting in 12 settings. In latent space, we use FM with LightningDiT on ImageNet, including class-conditional DiT-B/L/XL and unconditional DiT-B/L. The multi-shot extension is evaluated on six CIFAR settings with \texttt{UNet\,(Ho)} and two DiT settings.

\textbf{Implementation.} Within each setting, the baseline and our variants use matched training/evaluation protocols and random seeds, and differ only in the zero-initialized feedback branch. We only use horizontal flipping for data augmentation. Our implementations are primarily based on DDAE \cite{ddae} and LightningDiT \cite{lightning}. For UNets, since the formulations were introduced in separate lines of work with different training recipes and network details, we re-implement them in a unified codebase with shared training and evaluation pipelines across formulations and backbones. The main results therefore compare our models with their corresponding matched baselines, rather than with numbers reported in the original papers. CIFAR models are trained for 2000 epochs; ImageNet DiTs are trained for 100 epochs in class-conditional settings and 400 epochs in unconditional settings. All experiments are conducted with at most six NVIDIA RTX 4090 GPUs.

\begin{figure*}[t!]
    \centering
    \includegraphics[width=0.93\linewidth]{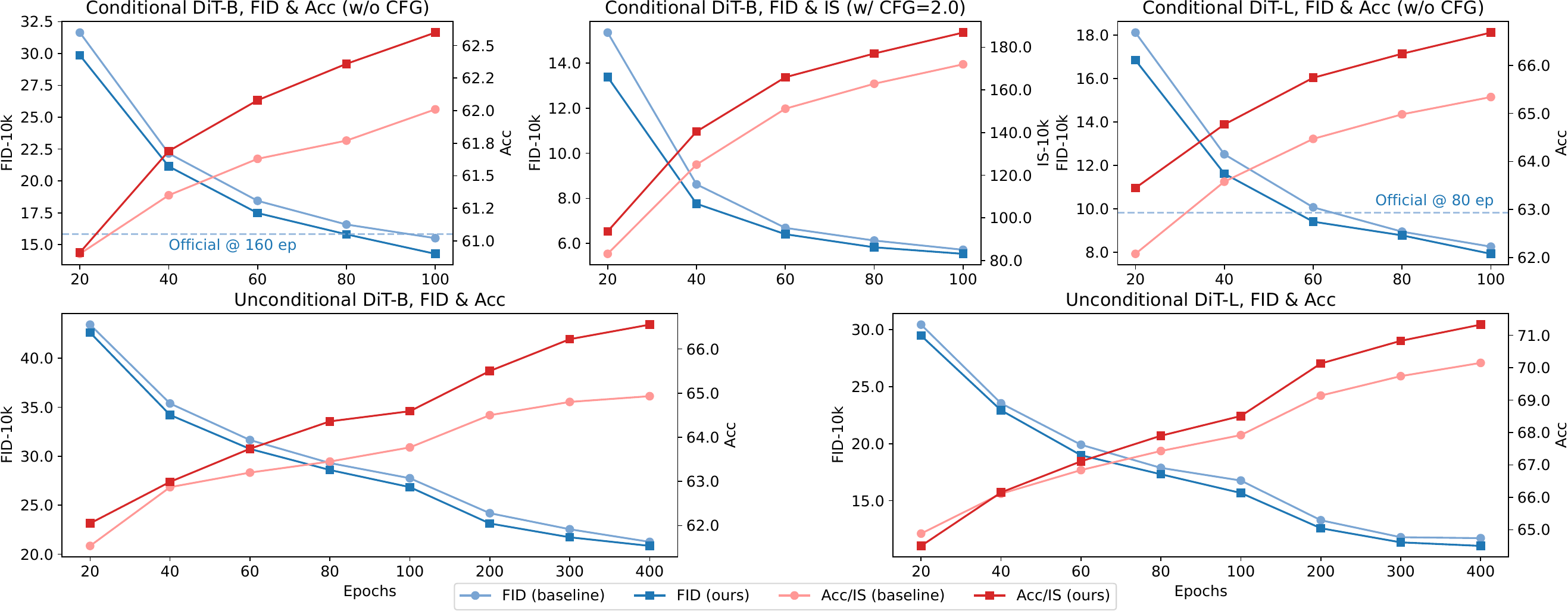}
    \caption{
    \textbf{Evolution of Generative and Discriminative Performance on ImageNet.}
    We compare matched LightningDiT baselines and our single-shot variants on class-conditional and unconditional DiT-B/L. The panels report FID and linear probing accuracy; for class-conditional DiT-B, we also report FID and IS with classifier-free guidance (CFG) \cite{cfg}.
    Conditioning Residuals generally achieve lower FID and higher accuracy/IS over training, with the accuracy advantage continuously widening in most settings.
    FID and IS are computed from 10K generated samples in this figure for efficient monitoring.
    }
    \label{fig:imagenet}
\end{figure*}

\textbf{Metrics.} We report FID \cite{fid} (computed from 50K samples by default) for all generation experiments, with IS \cite{inception} added in Fig.~\ref{fig:imagenet}. Representation quality is measured by linear probing accuracy on frozen backbones \cite{ddae, repa}, using the null class embedding for conditional models. For unconditional DiTs trained on ImageNet, we also evaluate dense transfer~using semantic segmentation mIoU. More evaluation details are in the supplement. Mechanistic analyses include layer-wise accuracy, CKA similarity \cite{cka}, training loss, and feature magnitude.

% ==================== Table 2: Baseline + Single ====================
\begin{table*}[t!]
    \centering
    \caption{A Comprehensive CIFAR Comparison of Generative and Discriminative Performance}
    \label{tab:baseline_single}
\begin{minipage}{0.49\textwidth}
\centering
\resizebox{\textwidth}{!}{
\begin{tabular}{llllll}
\hline
\multicolumn{6}{l}{\textbf{Unconditional CIFAR-10 Generation \& Linear Probing}} \\ \hline
\multirow{2}{*}{Model} & \multicolumn{1}{l|}{\multirow{2}{*}{Backbone}} & \multicolumn{2}{c|}{\textbf{Baseline}}     & \multicolumn{2}{c}{\textbf{+ Conditioning Residuals}} \\
        & \multicolumn{1}{l|}{}        & FID$\downarrow$ & \multicolumn{1}{l|}{Acc.$\uparrow$}  & FID$\downarrow$ & Acc.$\uparrow$ \\ \hline
DDPM    & \multicolumn{1}{l|}{UNet (Ho)}    & 3.52 & \multicolumn{1}{l|}{90.28} & 3.45 \textcolor{darkgreen}{(-0.07, 2.0\%)} & 90.75 \textcolor{darkgreen}{(+0.47)} \\
EDM     & \multicolumn{1}{l|}{UNet (Ho)}    & 3.54 & \multicolumn{1}{l|}{90.78} & 3.25 \textcolor{darkgreen}{(-0.29, 8.2\%)} & 91.24 \textcolor{darkgreen}{(+0.46)} \\
FM      & \multicolumn{1}{l|}{UNet (Ho)}    & 3.94 & \multicolumn{1}{l|}{90.43} & 3.74 \textcolor{darkgreen}{(-0.20, 5.1\%)} & 90.72 \textcolor{darkgreen}{(+0.29)} \\ \hline
DDPM    & \multicolumn{1}{l|}{UNet (Song)}  & 2.92 & \multicolumn{1}{l|}{93.81} & 2.79 \textcolor{darkgreen}{(-0.13, 4.5\%)} & 94.13 \textcolor{darkgreen}{(+0.32)} \\
EDM     & \multicolumn{1}{l|}{UNet (Song)}  & 2.23 & \multicolumn{1}{l|}{94.87} & 2.12 \textcolor{darkgreen}{(-0.11, 4.9\%)} & 94.76 \textcolor{darkred}{(-0.11)}   \\
FM      & \multicolumn{1}{l|}{UNet (Song)}  & 2.62 & \multicolumn{1}{l|}{93.98} & 2.52 \textcolor{darkgreen}{(-0.10, 3.8\%)} & 94.01 \textcolor{darkgreen}{(+0.03)} \\ \hline
\end{tabular}
}
\end{minipage}
\begin{minipage}{0.49\textwidth}
\centering
\resizebox{\textwidth}{!}{
\begin{tabular}{llllll}
\hline
\multicolumn{6}{l}{\textbf{Unconditional CIFAR-100 Generation \& Linear Probing}} \\ \hline
\multirow{2}{*}{Model} & \multicolumn{1}{l|}{\multirow{2}{*}{Backbone}} & \multicolumn{2}{c|}{\textbf{Baseline}}     & \multicolumn{2}{c}{\textbf{+ Conditioning Residuals}} \\
        & \multicolumn{1}{l|}{}        & FID$\downarrow$ & \multicolumn{1}{l|}{Acc.$\uparrow$}  & FID$\downarrow$ & Acc.$\uparrow$ \\ \hline
DDPM    & \multicolumn{1}{l|}{UNet (Ho)}    & 5.72 & \multicolumn{1}{l|}{62.07} & 5.67 \textcolor{darkgreen}{(-0.05, 0.9\%)} & 63.10 \textcolor{darkgreen}{(+1.03)} \\
EDM     & \multicolumn{1}{l|}{UNet (Ho)}    & 6.01 & \multicolumn{1}{l|}{63.65} & 5.82 \textcolor{darkgreen}{(-0.19, 3.2\%)} & 64.55 \textcolor{darkgreen}{(+0.90)} \\
FM      & \multicolumn{1}{l|}{UNet (Ho)}    & 6.48 & \multicolumn{1}{l|}{60.82} & 5.96 \textcolor{darkgreen}{(-0.52, 8.0\%)} & 62.22 \textcolor{darkgreen}{(+1.40)} \\ \hline
DDPM    & \multicolumn{1}{l|}{UNet (Song)}  & 4.40 & \multicolumn{1}{l|}{69.23} & 4.06 \textcolor{darkgreen}{(-0.34, 7.7\%)} & 70.46 \textcolor{darkgreen}{(+1.23)} \\
EDM     & \multicolumn{1}{l|}{UNet (Song)}  & 3.48 & \multicolumn{1}{l|}{70.16} & 3.37 \textcolor{darkgreen}{(-0.11, 3.2\%)} & 71.38 \textcolor{darkgreen}{(+1.22)} \\
FM      & \multicolumn{1}{l|}{UNet (Song)}  & 4.16 & \multicolumn{1}{l|}{68.16} & 3.90 \textcolor{darkgreen}{(-0.26, 6.3\%)} & 69.40 \textcolor{darkgreen}{(+1.24)} \\ \hline
\end{tabular}
}
\end{minipage}

\vspace{1.5ex}
\begin{minipage}{0.95\textwidth}
\textbf{Baseline:} standardized re-implementations over 12 settings with three formulations, two UNets, and two datasets. \textbf{+ Conditioning Residuals:} \textit{single-shot feedback} is added as the only architectural change. Parentheses report absolute changes and relative reductions for FID, and absolute point changes for Acc. Colored deltas denote \textcolor{darkgreen}{improvement} or \textcolor{darkred}{degradation}. Detailed results and a three-seed robustness analysis are reported in supplementary Tab.~\ref{tab:cifar_seeds}.

\end{minipage}
\end{table*}
% ==================== Table 2: Baseline + Single ====================

\subsection{Single-shot Evaluation on Pixel-space UNets}
Table~\ref{tab:baseline_single} provides a comprehensive controlled evaluation of our primary configuration. The purpose is to isolate whether a minimal feedback branch can broadly couple dual gains in generation and representation, while all other factors are matched across 12 settings and 24 evaluations.

\textbf{Generation} results show FID reductions in all 12 settings, with an average relative reduction of about 5\%. The gains are observed across both lighter and stronger UNet backbones and all three formulations, with only 0.46M extra parameters. This consistency shows that the effect is not tied to a specific prediction target, objective, sampler, or UNet design, and instead persists across models sharing the conditioning interface. Supplementary FID results further show that the improvements generally persist across training epochs rather than appearing only at the final reported checkpoints. Three-seed CIFAR-100 results with \texttt{UNet\,(Ho)} further support \textit{robustness}, with the smaller gains in Table~\ref{tab:baseline_single} even more pronounced on average.

\textbf{Linear probing} accuracy also improves in 11 cases, with only one minor exception. The gains are particularly evident on the more diverse CIFAR-100, where all six settings improve by around 1.2 points on average. As a reference, contrastive baselines reported in \cite{solo} typically achieve roughly 89--93\% on CIFAR-10 and 64--72\% on CIFAR-100. Many of our results already fall into this range, even though they are trained only with diffusion objectives without any discriminative designs.

These paired improvements are achieved concurrently under the same minimal modification. The feedback path that makes encoded semantics available to decoding layers improves FID, while the same compact route also encourages these features to become more reusable and linearly separable. Having verified this, we next examine whether the same mechanism remains effective, scalable, and transferable in modern ImageNet DiTs.

% ==================== Figure 3: Epochs on ImageNet ====================
% ==================== Figure 3: Epochs on ImageNet ====================

\subsection{Single-shot Evaluation on Latent-space DiTs}
Following LightningDiT and its VA-VAE latent space \cite{lightning}, we move to the larger-scale ImageNet at $256{\times}256$ resolution. The evaluation is performed on both class-conditional and unconditional models. For class-conditional DiT-B/L/XL, we mainly examine whether our method can further improve the already strong generation quality. For unconditional DiT-B/L, we focus more on representation quality, since this setting corresponds to a standard self-supervised learning scenario and usually yields better features than conditional models \cite{ldae}. In both regimes, single-shot feedback is the only change.

\textbf{Scaling behavior} is demonstrated in Fig.~\ref{fig:imagenet}, which traces the detailed evolution of generative and discriminative performance. The first row shows class-conditional models scaling from DiT-B to DiT-L, while the second row shows unconditional models scaling by model size and training epoch. On top of this already optimized baseline, our method yields a consistent FID/IS advantage throughout training, indicating that the feedback branch remains effective beyond early convergence. The accuracy advantage is also persistent and often widens as training proceeds. In class-conditional models, our accuracy around 60 epochs surpasses the baseline at 100 epochs; in unconditional models, our accuracy at 200 epochs already surpasses the baseline at 400 epochs. This suggests up to nearly $2\times$ faster representation learning and higher final accuracy. More importantly, representation feedback does not merely improve one metric at the expense of the other, but links both through the reused pathway.

% ==================== Table 3: Comparison on ImageNet ====================
\begin{table}[t]
    \centering
    \caption{ImageNet Generation and Representation Comparison}
    \label{tab:imagenet}
    \resizebox{0.9\linewidth}{!}{
    \begin{tabular}{@{}lccrr@{\hspace{10pt}}r@{}}
        \toprule
        \multirow{2}{*}{\textbf{Model}} & \multirow{2}{*}{\textbf{Backbone}} & \multirow{2}{*}{\textbf{Rep. setup}} & \multicolumn{3}{c}{\textbf{ImageNet}} \\
        \cmidrule(lr){4-6}
                  &       &     & {Epoch} & {FID$^\dagger$$\downarrow$} & {LP Acc.$\uparrow$} \\
        \midrule
        \multicolumn{4}{l}{\textcolor{gray}{Flattened, clean VA-VAE latent code}} & \textcolor{gray}{\NA} & \textcolor{gray}{40.04} \\
        \midrule
        LightningDiT (u) & DiT-B & \NA & 400 & 18.86 & 64.93 \\
        \rowcolor{pinktab}
        + ours    & DiT-B & \NA & 400 & 17.94 & 66.55 \\
        \midrule
        LightningDiT (u) & DiT-L & \NA & 400 &  8.73 & 70.15 \\
        \rowcolor{pinktab}
        + ours    & DiT-L & \NA & 400 &  8.07 & 71.33 \\
        \midrule
        LightningDiT (c) & DiT-B & \NA & 100 &  2.87 / 12.70 & 62.01 \\
        \rowcolor{pinktab}
        + ours    & DiT-B & \NA & 100 &  2.67 / 11.89 & 62.60 \\
        \midrule
        LightningDiT (c) & DiT-L & \NA & 100 &   2.09 / 5.53 & 65.34 \\
        \rowcolor{pinktab}
        + ours    & DiT-L & \NA & 100 &   2.04 / 5.30 & 66.68 \\
        \midrule
        LightningDiT (c) & DiT-XL& \NA & 100 &   1.79 / 4.72 & 66.74 \\
        \rowcolor{pinktab}
        + ours    & DiT-XL& \NA & 100 & \textbf{1.73 / 4.52} & 67.15 \\
        \midrule
        \multirow{3}{37pt}{REPA (cond) \cite{repa}} & DiT-B & DINOv2* & 80 & 24.40 & 61.20 \\
                                                    & DiT-L & DINOv2* & 80 & 10.00 & 69.40 \\
                                                    & DiT-XL& DINOv2* & 80 &  7.90 & 70.30 \\
        \midrule
        \multirow{2}{37pt}{\textit{l}-DAE \cite{ldae}} & DiT-L & \NA       & 400 & 11.60 & 57.50 \\
                                                       & DiT-L & rand crop & 400 & \NA   & 65.00 \\
        \midrule
        \textcolor{gray}{DINO \cite{dino}}            & \textcolor{gray}{ViT-B} & \textcolor{gray}{CL} & \textcolor{gray}{400} & \textcolor{gray}{\NA} & \textcolor{gray}{\textbf{78.20}} \\
        \textcolor{gray}{DINO$^\#$\,(in \cite{soda})} & \textcolor{gray}{ViT-B} & \textcolor{gray}{CL} & \textcolor{gray}{400} & \textcolor{gray}{\NA} & \textcolor{gray}{65.30} \\
        \midrule
        \textcolor{gray}{MAE (in \cite{SGMAE})} & \textcolor{gray}{ViT-B} & \textcolor{gray}{MIM} & \textcolor{gray}{400} & \textcolor{gray}{\NA} & \textcolor{gray}{61.40} \\
        \textcolor{gray}{MAE \cite{mae}}        & \textcolor{gray}{ViT-L} & \textcolor{gray}{MIM} & \textcolor{gray}{400} & \textcolor{gray}{\NA} & \textcolor{gray}{69.70} \\
        \bottomrule
    \end{tabular}
    }

\vspace{1.5ex}
\begin{minipage}{0.95\linewidth}
\textbf{Rep. setup} summarizes representation-oriented designs during training, including data augmentation beyond flipping, self-supervised paradigms such as contrastive learning (CL) and masked image modeling (MIM), or external representation alignment.
\textbf{(u) and (c)} denote unconditional and class-conditional models.
The probe contains a parameter-free BatchNorm and a linear layer, following \textit{l}-DAE.
$^\dagger$For LightningDiT (c), $a$ / $b$ denotes FID with CFG and without CFG, respectively; others are reported without CFG.
$^*$REPA aligns DiT hidden states to an external DINOv2 encoder.
$^\#$DINO using only flip and crop, with stronger augmentations removed.
\end{minipage}
\end{table}
% ==================== Table 3: Comparison on ImageNet ====================

\textbf{Final results} are summarized in Table~\ref{tab:imagenet}. In the unconditional setting, Conditioning Residuals improve both generation and representation, reducing FID by 0.92/0.66 and increasing accuracy by 1.62/1.18 points on DiT-B/L, respectively. The same trend holds for conditional models, where our single-shot approach consistently improves FID both with and without classifier-free guidance (CFG) \cite{cfg}, as well as accuracy. Notably, our LightningDiT baseline, alongside REPA, already represents a leading tier of highly optimized approaches with extremely fast convergence. On top of this strong baseline, our minimal modification yields further gains, reaching FID 1.73 and 4.52 on DiT-XL. Together with Fig.~\ref{fig:imagenet}, these results show that the dual improvements are consistent and scalable under diverse conditioning and sampling settings. Generated visual samples are presented in the supplementary material.

The remaining rows in Table~\ref{tab:imagenet} provide references across related methods. \textit{l}-DAE \cite{ldae} and MAE \cite{mae} are also based on the broad denoising-autoencoding paradigm, but they are designed for representation learning rather than full generative modeling. Relative to these references, our unconditional DiT-B/L achieve higher linear probing accuracy while retaining full generative capability.
Dedicated representation learners such as DINO \cite{dino} are also included, though they rely on objectives and augmentation pipelines explicitly designed for learning better encodings. While the full DINO result remains higher, our diffusion-native models already numerically surpass the DINO variant without strong augmentations.
REPA \cite{repa} represents another route: it directly aligns DiT hidden states with a strong DINOv2 encoder. In contrast, our results demonstrate that even without an external representation component, making the model's own intermediate features actively reusable as conditioning can still improve generation quality.

% ==================== Table 4: Segmentation ====================
\begin{table}[t]
    \centering
    \caption{Dense Linear Probing for Semantic Segmentation}
    \label{tab:segmentation}
    \resizebox{0.5\linewidth}{!}{
    \begin{tabular}{@{}lcc@{}}
        \toprule
        \textbf{Model /} & \textbf{ADE20K} & \textbf{VOC2012} \\
        \cmidrule(l){2-2} \cmidrule(l){3-3}
        \textbf{Pre-train epoch}  & {mIoU$\uparrow$} & {mIoU$\uparrow$} \\
        \midrule
        DiT-B, 100 ep.  & 29.75  & 62.53 \\
        \rowcolor{pinktab}
        + ours    & 30.40  & 63.26 \\
        \midrule
        DiT-B, 200 ep.  & 30.07  & 63.45 \\
        \rowcolor{pinktab}
        + ours    & 30.72  & 64.33 \\
        \midrule
        DiT-B, 400 ep.  & 30.48  & 64.34 \\
        \rowcolor{pinktab}
        + ours    & \textbf{30.75}  & \textbf{65.61} \\
        \midrule
        \textcolor{gray}{ViT-B/8 \cite{dinov2}} & \multirow{2}{*}{\textcolor{gray}{31.80}} & \multirow{2}{*}{\textcolor{gray}{66.40}} \\
        \textcolor{gray}{(DINO, $512^2$)} & & \\
        \midrule
        DiT-L, 400 ep.  & 33.04  & 70.52 \\
        \rowcolor{pinktab}
        + ours    & \textbf{33.77}  & \textbf{71.34} \\
        \bottomrule
    \end{tabular}
    }

\vspace{1.5ex}
\begin{minipage}{0.95\linewidth}
Following \cite{ibot}, we concatenate frozen patch tokens ($16^2$) from multiple layers to leverage multi-level features: layers 5 and 7 for DiT-B, and layers 9, 11, 13, and 15 for DiT-L. These layers are fixed for both baselines and our models. The prediction head contains a LayerNorm and a linear layer, trained without data augmentation. Models are LightningDiT (u).
\end{minipage}
\end{table}
% ==================== Table 4: Segmentation ====================

\textbf{Transfer learning on semantic segmentation.} Conditioning Residuals also enhance local features transferable to dense tasks. We validate this by dense linear probing on ADE20K \cite{ade20k} and VOC2012 \cite{voc2012}, where the backbone is frozen and only a linear head is trained to predict segmentation labels from a coarse $16\times16$ patch token grid. As shown in Table~\ref{tab:segmentation}, our method consistently outperforms the baselines across datasets, model sizes, and pre-training epochs. Our 100-epoch DiT-B model surpasses the 200-epoch baseline on ADE20K, and our 200-epoch model surpasses the 400-epoch baseline; on VOC2012, our gain widens from 0.73 to 1.27 mIoU as pre-training proceeds. The same trend holds when scaling to DiT-L, with the DiT-B/L results broadly competitive with the DINO ViT-B/8 reference reported in \cite{dinov2}. These results indicate that the feedback's benefits extend from global representations to dense content understanding, where the advantage appears earlier and reaches higher final performance.

\subsection{Generative Performance of Multi-shot Feedback}

% ==================== Figure 4: Multi UNet ====================
\begin{figure}[t]
\centering
    \includegraphics[width=\linewidth]{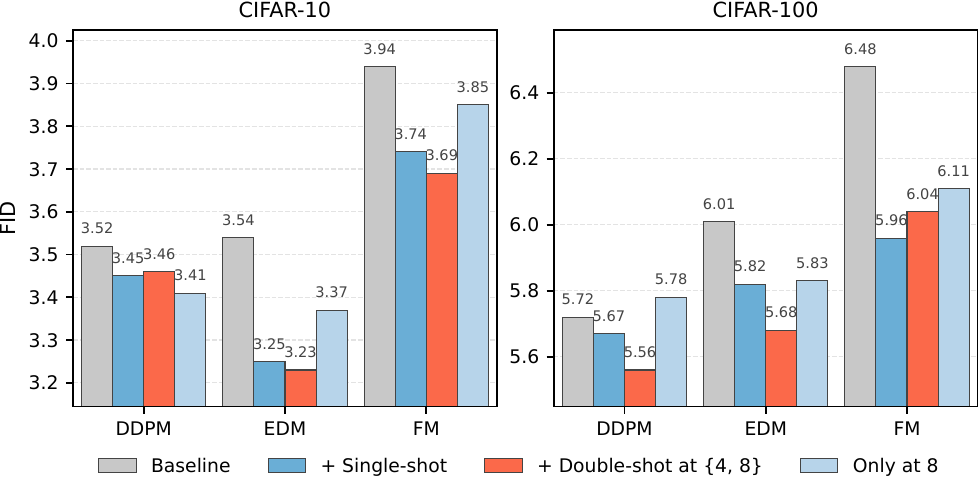}
\caption{
    \textbf{Multi-shot Generative Performance on CIFAR.}
    We compare the double-shot variant $\mathcal{M}=\{4,8\}$ with the matched baseline, the main single-shot configurations from Tab.~\ref{tab:baseline_single}, and a control setup that only uses layer 8.
}
\label{fig:multi}
\end{figure}
% ==================== Figure 4: Multi UNet ====================

% ==================== Table 5: Multi DiT ====================
\begin{table}[t]
    \centering
    \caption{Multi-shot Feedback on ImageNet DiTs}
    \label{tab:multi_dit}
    \resizebox{0.9\linewidth}{!}{
    \begin{tabular}{@{}llccc@{}}
        \toprule
        \multirow{2}{*}{\textbf{Backbone}} & \multirow{2}{*}{\textbf{Method}} & \multirow{2}{*}{\textbf{Projection}} & \multicolumn{2}{c}{\textbf{ImageNet}} \\
        \cmidrule(lr){4-5}
                  &       &     & {FID w/ CFG$\downarrow$} & {FID w/o CFG$\downarrow$} \\
        \midrule
        DiT-B & Baseline        & \NA      & 2.87 & 12.70 \\
        DiT-B & +\,Single-shot  & \NA      & 2.67 & 11.89 \\
        DiT-B & +\,Double-shot  & Separate & 2.69 & 11.71 \\
        \rowcolor{pinktab}
        DiT-B & +\,Double-shot  & Shared   & \textbf{2.62} & \textbf{11.58} \\
        \midrule
        DiT-L & Baseline        & \NA      & 2.09 & 5.53 \\
        DiT-L & +\,Single-shot  & \NA      & 2.04 & 5.30 \\
        DiT-L & +\,Double-shot  & Separate & 2.03 & \textbf{5.28} \\
        \rowcolor{pinktab}
        DiT-L & +\,Double-shot  & Shared   & \textbf{2.01} & 5.34 \\
        \bottomrule
    \end{tabular}
    }
    
\vspace{1.5ex}
\begin{minipage}{0.95\linewidth}
Models are LightningDiT (c) under the same protocol as Tab.~\ref{tab:imagenet}. Baseline and single-shot results are copied there. Shared projection follows Eq.~(\ref{eq:shot_shared}).
\end{minipage}
\end{table}
% ==================== Table 5: Multi DiT ====================

The single-shot method treats one selected layer as a semantic snapshot that is ready to condition subsequent denoising. However, a single layer near a strong probing peak may not capture the whole useful semantic region, since earlier layers may still provide complementary information. We therefore evaluate whether generation benefits from repeated feedback, rather than from simply tuning the position.

For \texttt{UNet\,(Ho)}, we use the fixed double-shot schedule $\mathcal{M}=\{4,8\}$ and add a control that only uses layer 8 and removes the earlier feedback point. Across all six settings in Fig.~\ref{fig:multi}, double-shot feedback improves FID over the baseline. More importantly, it outperforms the layer-8 control in five cases, with only a small exception, showing that the earlier point contributes non-redundant generative guidance beyond a later semantic feature. Compared with the main single-shot results in Table~\ref{tab:baseline_single}, double-shot feedback remains competitive and often better. This matches its intended role: a fixed, generation-oriented extension that can extract additional gains without per-setting search, while leaving single-shot feedback as the main approach for dual gains in generation and representation.

Table~\ref{tab:multi_dit} further evaluates the same idea on ImageNet. On DiT-B, double-shot feedback with shared projection improves FID over the main single-shot results both with and without CFG. This design in Eq.~(\ref{eq:shot_shared}) also outperforms the variant with separate projections, suggesting that mapping summaries from different layers into a common residual space is beneficial. On DiT-L, double-shot feedback still improves over the baseline and further improves the CFG FID from single-shot's 2.04 to 2.01. These results indicate that repeated feedback is also useful for generative performance in modern latent-space DiTs. Qualitative comparisons in the supplementary material visualize the corresponding DiT-L improvements.

\subsection{Mechanistic Analysis}

\begin{figure}[t!]
\centering
\includegraphics[width=\linewidth]{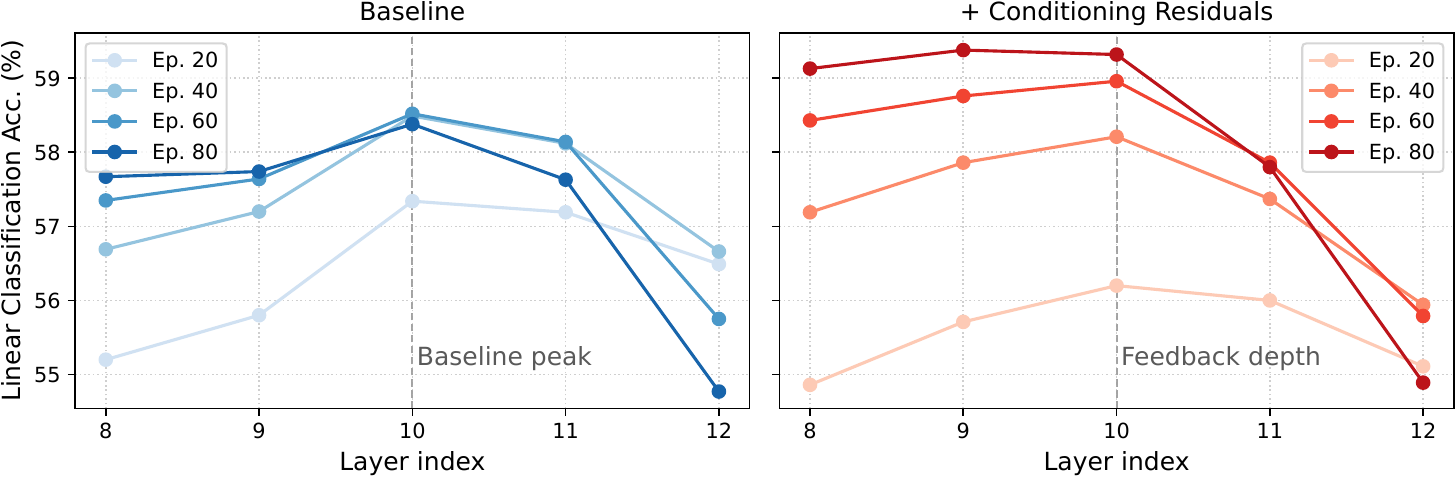}
\caption{
    \textbf{Layer-wise linear classification as training progresses.}
    \textit{Left:} In the baseline, linear separability naturally peaks around layer 10, with high values spread over layers 8--11.
    \textit{Right:} For this diagnostic only, feedback is placed at the baseline peak to isolate its effect. The high-separability region shifts earlier to layers 8--10 and rises higher, with a larger drop toward near-output layers, forming a clearer hierarchy.
    A single linear layer is trained to examine raw linear separability. Models are class-conditional 12-layer LightningDiT-B.
}
\label{fig:layerwise}
\end{figure}

\textbf{Layer-wise linear separability} is first used as a diagnostic of how feedback reshapes the distribution of semantics formed in generative training. Fig.~\ref{fig:layerwise} shows classification results from different layers. In the baseline, discriminative power naturally emerges near layer 10, but remains distributed across a late region, and the gains largely plateau. This suggests that diffusion training recovers useful semantics, but does not explicitly concentrate them toward earlier layers before prediction.

To investigate the effect of single-shot feedback, we place it at this peak layer. With feedback, the peak accuracy continues to rise, and the semantically rich region separates more clearly from late, near-output layers, consistent with a stronger early semantic bottleneck and a more pronounced layer-wise hierarchy. This shows that the feedback mechanism does not merely read out existing information, but actively reorganizes reusable semantics before returning them as denoising conditions.

Interestingly, the peak layer after feedback does not always coincide exactly with the feedback depth. In this case, the peak appears slightly before the feedback is injected. This indicates that linear separability and usefulness for conditioning are related but not identical, a gap we further analyze below.

\begin{figure}[t]
\centering
    \includegraphics[width=\linewidth]{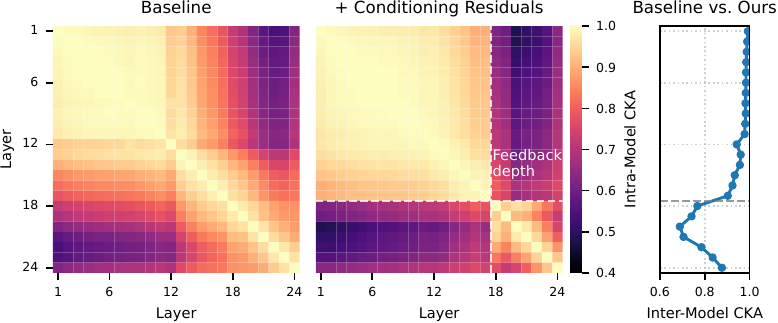}
\caption{
    \textbf{CKA analysis of representation structure.}
    \textit{Left} and \textit{middle} panels show intra-model CKA for the baseline and our model.
    The \textit{right} panel reports inter-model CKA between their corresponding layers.
    Our method induces a clearer bottleneck-like structural transition around the feedback depth.
}
\label{fig:cka}
\end{figure}

\textbf{CKA representation structure.}
In Fig.~\ref{fig:cka}, we analyze the feature similarity structure using Centered Kernel Alignment (CKA) \cite{cka}. For clearer visualization, we use the 24-layer DiT-L. In the baseline, the intra-model similarity exhibits a smooth and gradual decay as depth increases. Although there appears to be a weak structural change, the transitions between adjacent layers remain smooth, and there is no clear boundary separating a semantic accumulation phase from a decoding phase. This continuous spectrum suggests that the baseline performs generation through a gradual evolution of features, with encoding and decoding roles mostly entangled.

% ==================== Figure 5: Layer-wise Evolution ====================
% ==================== Figure 5: Layer-wise Evolution ====================

% ==================== Figure 6: CKA ====================
% ==================== Figure 6: CKA ====================

With feedback at layer 17, the CKA map exhibits a more pronounced bottleneck-like structural transition. Layers 1--17 remain mutually similar, whereas their similarity to the subsequent layers decreases sharply, and layers 18--24 show lower off-diagonal similarity than in the baseline. Since these later layers are explicitly modulated by the feature summary, their hidden states need not rely solely on layer-wise propagation to preserve recovered information. This additional route may allow these layers to diverge more rapidly and adapt to the parameterization-specific prediction target near the output end.

Fig.~\ref{fig:cka} also explains why the probing peak, layer 13 in this case, can precede the feedback depth. Within the coarse pre-feedback stage, there is a finer transition: layers 13--17 remain mutually similar but are less aligned with layers 1--12, and start to diverge from the baseline after inter-model CKA remains close to one through layer 11.
These patterns suggest that the most discriminative feature is not immediately read out, but continues to evolve over a short late-encoding adaptation stage. Thus, the feedback depth marks the end of this adaptation and a suitable readout point for generative guidance.

Finally, the inter-model comparison shows that their divergence becomes substantially stronger after the feedback depth. This indicates that Conditioning Residuals intervene in the output stage, reshaping these layers for specialized decoding. Together, these intra-model and inter-model observations support a more decoupled internal structure, where the added conditioning route complements direct layer-wise propagation.

% ==================== Figure 7: Loss ====================
\begin{figure}[t]
\centering
    \includegraphics[width=\linewidth]{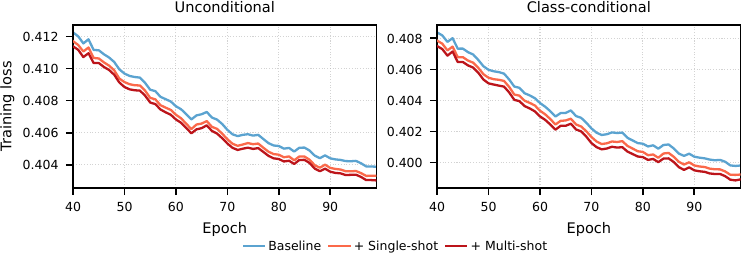}
\caption{
    \textbf{Training loss trajectories.}
    Both variants of Conditioning Residuals facilitate optimization and reduce training loss under both unconditional and class-conditional training.
    Curves are smoothed for visualization. Models are LightningDiT-B trained with the official composite objective.
}
\label{fig:loss}
\end{figure}
% ==================== Figure 7: Loss ====================

\textbf{Training loss dynamics.}
We next examine whether this feature structure facilitates model optimization. Fig.~\ref{fig:loss} compares the baselines with our single-shot and multi-shot variants. Unlike FID or probing, the loss directly reflects the native diffusion training status. In both settings, Conditioning Residuals lead to consistently lower loss trajectories, with multi-shot feedback providing a further reduction. Together with the gains in generation and representation, these lower loss trajectories suggest that the improved information flow guides training toward a better optimization outcome, beyond post-hoc evaluation metrics for specific tasks.

% ==================== Figure 8: Magnitude ====================
\begin{figure}[t]
\centering
    \includegraphics[width=\linewidth]{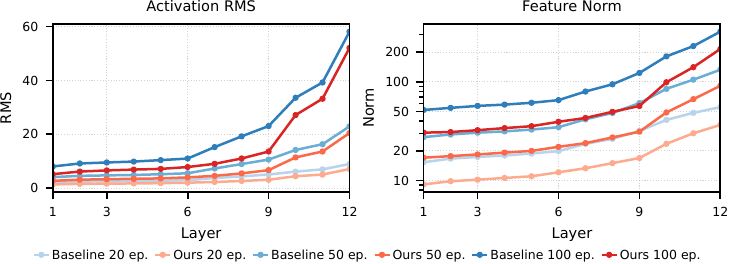}
\caption{
    \textbf{Layer-wise magnitude profiles as training progresses.}
    \textit{Left:} Root Mean Square (RMS) of raw activation tensors.
    \textit{Right:} L2 norm of globally averaged features after mean-centering.
    Conditioning Residuals suppress the depth-wise and training-time growth of both quantities, suggesting more stable training dynamics without evident scale inflation.
    Models are class-conditional LightningDiT-B trained on ImageNet at 20, 50, and 100 epochs.
}
\label{fig:magnitude}
\end{figure}
% ==================== Figure 8: Magnitude ====================

\textbf{Feature magnitude.}
Motivated by magnitude-based analyses of diffusion training dynamics \cite{edm2}, Fig.~\ref{fig:magnitude} tracks how feature scales evolve across depth and training duration. Conditioning Residuals consistently lower the activation RMS relative to the baseline and control the norm of mean-centered global features. Since these reductions coincide with improved generation, stronger representations, and lower diffusion loss, the gains are not driven by spurious scale inflation. Instead, the feedback branch may act as a regularizer on intermediate states, limiting magnitude drift so that later denoising updates remain effective. This also provides a plausible explanation for the lower training loss trajectories, in line with the broader role of magnitude control in deep generative models \cite{dar, edm2}.

\subsection{Ablation Study}
We ablate two core design choices: how the feedback residual is formed, and how the feedback depth is selected.
Table~\ref{tab:ablation} studies timestep modulation in Eq.~(\ref{eq:shot}) on CIFAR-100. Time-dependent scaling gives the best FID across all three formulations, whereas direct addition does not consistently improve generation, though it may improve accuracy more in some cases. This suggests that timestep-aware feedback prioritizes generation in this trade-off by adapting the cue to the denoising trajectory rather than injecting it statically.

We also validate our layer selection strategy on ImageNet. Across five studied variants, early loss, final FID, and accuracy improve successively from the baseline to $m=8$ and then to $m=9$, and deteriorate through $m=10$ and $m=11$. The one with the lowest loss after 20 epochs also achieves the best FID and accuracy at 100 epochs. This aligned trend suggests that early diffusion loss is predictive of final relative performance across both metrics without full training. This relation is consistent with Fig.~\ref{fig:loss} and the view that improved information flow is reflected in better training dynamics.

% ==================== Table 6: Ablation ====================
\begin{table}[t]
    \centering
    \caption{Ablation Studies on Feedback Head and Placement}
    \label{tab:ablation}
    \scriptsize
    \setlength{\tabcolsep}{4.3pt}

    \subfloat[Time modulation on UNet (Ho)]{
    \begin{tabular}{@{}llcc@{}}
        \hline
        Model & Variant & FID$\downarrow$ & Acc.$\uparrow$ \\
        \hline
        DDPM & Baseline & 5.72 & 62.07 \\
        DDPM & Direct add. & 5.70 & \textbf{63.56} \\
        \rowcolor{ablatab}
        DDPM & Time-depen. & \textbf{5.67} & 63.10 \\
        \hline
        EDM & Baseline & 6.01 & 63.65 \\
        EDM & Direct add. & 6.14 & \textbf{64.84} \\
        \rowcolor{ablatab}
        EDM & Time-depen. & \textbf{5.82} & 64.55 \\
        \hline
        FM & Baseline & 6.48 & 60.82 \\
        FM & Direct add. & 6.25 & 61.77 \\
        \rowcolor{ablatab}
        FM & Time-depen. & \textbf{5.96} & \textbf{62.22} \\
        \hline
    \end{tabular}
    }
    \hfill
    \subfloat[Layer selection with profiling runs on class-conditional LightningDiT-B]{
    \begin{tabular}{@{}cccc@{}}
        \hline
        Variant & Loss$\downarrow$ & FID-10K$\downarrow$ & Acc.$\uparrow$ \\
        & 20 ep. & 100 ep. & 100 ep. \\
        \hline
        Baseline & 0.4141 & 15.51 & 62.01 \\
        $m=8$ & 0.4135 & 14.79 & 62.47 \\
        \rowcolor{ablatab}
        $m=9$ & \textbf{0.4133} & \textbf{14.28} & \textbf{62.60} \\
        $m=10$& 0.4136 & 14.52 & 62.39 \\
        $m=11$& 0.4140 & 14.77 & 62.19 \\
        \hline
    \end{tabular}
    }
\end{table}
% ==================== Table 6: Ablation ====================

\section{Conclusion}
\noindent
This paper investigates an underexplored connection between representations and conditioning in diffusion models. While such internal states are often used for post-hoc evaluation, we examine whether they can serve as self-guidance signals to improve generative training itself. To explore this idea, we propose Conditioning Residuals, a lightweight mechanism that aggregates intermediate features and feeds them back as residuals through the native conditioning pathway in UNet and DiT. Without extra encoders, auxiliary objectives, or sampling-time changes, this simple architectural modification consistently improves generation quality in 17 pixel- and latent-space settings and generally strengthens learned representations, as shown by FID, linear classification, semantic segmentation, and qualitative results. Importantly, the implementation brings only negligible parameter overhead while leaving wall-clock cost essentially unchanged. The multi-shot extension further shows that repeated updates can provide additional generative gains, while mechanistic analyses reveal reshaped layer-wise feature structure and improved training dynamics.
These findings position representation feedback as a self-contained, generalizable mechanism for strengthening this shared generative foundation from within, and motivate its future extension to larger image, video, and multimodal generators.

\bibliographystyle{IEEEtran}
\bibliography{refs}

\clearpage

\appendices

% ==================== Figure 9: Visualization ====================
\begin{figure*}[t!]
    \centering
    \vspace{1.5em}
    {\LARGE Supplementary Material\par}
    \vspace{0.8em}
    {\Large for ``Conditioning Residuals for Diffusion Models via Representation Feedback''\par}
    \vspace{2\baselineskip}
    \includegraphics[width=\linewidth]{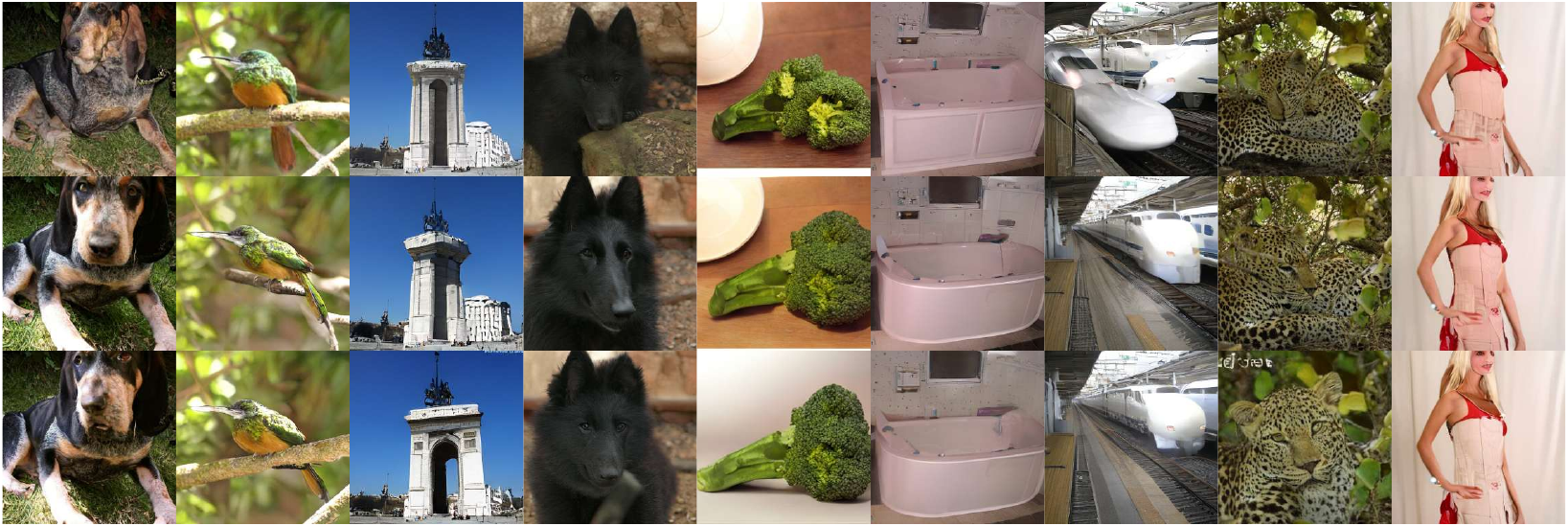}
    \caption{
    \textbf{Qualitative ImageNet samples.}
    We compare class-conditional LightningDiT-L models under matched class labels, initial noises, and guidance~settings.
    Rows correspond to the baseline (FID=2.09), single-shot feedback (FID=2.04), and double-shot feedback (FID=2.01), respectively.
    }
    \label{fig:visual}
    \vspace{1.8\baselineskip}
\end{figure*}
% ==================== Figure 9: Visualization ====================

\section{Qualitative Comparison}
\noindent
Fig.~\ref{fig:visual} provides generated ImageNet samples from the class-conditional LightningDiT-L baseline, the single-shot variant, and the double-shot variant. Samples in each column are matched by class label, initial noise, sampler and steps, and CFG settings. Under this controlled comparison, the feedback variants generally preserve the layout and, in several examples, produce more coherent object structures (\eg, clearer head and body in dog and leopard), more plausible global shapes (\eg, triumphal arch and bathtub), or better visual details (\eg, the broccoli). Although these changes do not fully resolve all failure cases, as expected from a lightweight architectural modification, the trends are broadly consistent with the quantitative improvements reported in the main paper.

\section{Wall-clock Overhead}
\noindent
Table~\ref{tab:overhead} reports parameter counts and wall-clock training and sampling overheads, using the UNet settings as representative cases. Single-shot feedback adds only 0.5M parameters and incurs minor sampling overhead. Double-shot feedback further adds another 0.5M and remains close to the baseline in wall-clock time. The training differences fall within the variation measured across the three formulations.

% ==================== Table 7: Overhead ====================
\begin{table}[t]
    \centering
    \caption{Parameter Count and Runtime Overhead on CIFAR-10}
    \label{tab:overhead}
    \resizebox{\linewidth}{!}{
    \begin{tabular}{@{}llccc@{}}
        \toprule
        \textbf{Backbone} & \textbf{Variant} & \textbf{Param. (M)} & \textbf{Train Over. (\%)} & \textbf{Sample Over. (\%)} \\
        \midrule
        UNet (Ho)   & Baseline    & 35.7 & \NA & \NA \\
        UNet (Ho)   & Single-shot & 36.2 & $+0.38\pm0.41$ & $+0.42\pm0.34$ \\
        UNet (Ho)   & Double-shot & 36.7 & $+0.52\pm0.78$ & $+1.12\pm0.44$\\
        \midrule
        UNet (Song) & Baseline    & 56.5 & \NA & \NA \\
        UNet (Song) & Single-shot & 57.0 & $+0.23\pm0.57$ & $-0.14\pm0.31$\\
        \bottomrule
    \end{tabular}
    }

\vspace{1.5ex}
\begin{minipage}{0.95\linewidth}
Training and sampling are measured as s/iter and s/1K images. Overheads are relative to matched baselines and reported as mean$\pm$std over the three DDPM, EDM, and FM formulations. FM uses the adaptive RK45 sampler.
\end{minipage}
\end{table}
% ==================== Table 7: Overhead ====================

\section{Training and Evaluation Protocol}
\noindent
Within each setting, all paired comparisons, \ie, the matched baseline and Conditioning Residuals variants, share the same training recipe, evaluation protocol, random seeds, and code. The feedback branch is the only architectural change.

Pixel-space DDPM, EDM, and FM are re-implemented in a unified codebase with standardized training/evaluation scripts and UNet block implementation details. Therefore, the reported results compare our models with their matched baselines, rather than with numbers separately reported from the original papers or obtained from different code repositories.

\subsection{Training}
On CIFAR, all models are trained for 2000 epochs. Checkpoints are evaluated every 200 epochs to report the best FID, which is typically achieved between 1400 and 2000 epochs. Linear probing is evaluated on a predefined set of checkpoints at 800, 1000, and 1200 epochs, and the best result is reported. The same checkpoint protocol is applied to all models. These experiments are conducted on two NVIDIA RTX 4090 GPUs.

\pagebreak[4]
On ImageNet, class-conditional models are trained for 100 epochs, and unconditional models are trained for 400 epochs. The best FID and linear probing results are consistently achieved at the final checkpoint and are reported in the main table. These experiments are conducted with six 4090 GPUs.

\subsection{Sampling}
For generative evaluation, each diffusion formulation follows the convention of its corresponding baseline. DDPM uses 100-step deterministic DDIM sampling, EDM uses the 18-step deterministic Heun solver, pixel-space FM uses the adaptive RK45, and latent-space FM follows LightningDiT with the 250-step Euler method. Within each paired comparison, the baseline and our variants share the same sampler, noise seed, noise schedule, and guidance setting when applicable.

% ==================== Table 8: Protocol ====================
\begin{table}[t]
\centering
\caption{Timestep and Layer Selection in Linear Probing}
\label{tab:protocol}
\begin{minipage}{0.56\linewidth}
\centering
\resizebox{\textwidth}{!}{
\begin{tabular}{@{}ll@{}}
\toprule
\multicolumn{2}{c}{\textbf{Timesteps}} \\
\midrule
Formulation & Search window \\
\midrule
DDPM, pixel & fixed $t=11$ \\
EDM, pixel  & steps $\{3,4\}$ of 18-step Heun \\
FM, pixel   & $t\in\{0.05,0.06,0.07\}$ \\
FM, latent  & fixed $t=0.25$ \\
\bottomrule
\end{tabular}
}
\end{minipage}
\begin{minipage}{0.42\linewidth}
\centering
\resizebox{\textwidth}{!}{
\begin{tabular}{@{}ll@{}}
\toprule
\multicolumn{2}{c}{\textbf{Layers}} \\
\midrule
Backbone & Semantic region \\
\midrule
UNet (Ho)   & output layers $6$--$8$ \\
UNet (Song) & output layers $7$--$9$ \\
DiT-B       & layers $6$--$10$ \\
DiT-L       & layers $8$--$10$, $13$--$16$ \\
DiT-XL      & layers $10$--$14$ \\
\bottomrule
\end{tabular}
}
\end{minipage}
\end{table}
% ==================== Table 8: Protocol ====================

% ==================== Table 9: Probe Details ====================
\begin{table}[t]
    \centering
    \caption{Representation Evaluation Details}
    \label{tab:probe_details}
    \resizebox{\linewidth}{!}{
    \begin{tabular}{@{}lllll@{}}
        \toprule
        \textbf{Experiment} & \textbf{Head} & \textbf{Epochs} & \textbf{Optimizer / Learning rate} & \textbf{Aug.} \\
        \midrule
        CIFAR LP     & linear      & 15 & Adam / 4e-3 & flip \\
        ImageNet LP  & BN + linear & 30 & Adam / 2e-3 (B), 1e-3 (L/XL) & flip \\
        Segmentation & LN + linear & 30 & Adam / 3e-3 & \NA \\
        \bottomrule
    \end{tabular}
    }
\end{table}
% ==================== Table 9: Probe Details ====================

\subsection{Representation Evaluation}
Linear probing is used as a post-hoc evaluation of frozen representations. Since diffusion representations are distributed across noise levels and network depths, the best combination of timestep and layer may vary across trained models. Following common practice, we use a factorized search window, as summarized in Table~\ref{tab:protocol}. The timestep window is tied to the diffusion formulation and is largely shared across backbones, while the layer region is tied to the backbone and is largely shared across formulations. For each trained model, including both baselines and our variants, we evaluate the corresponding timestep-layer candidates and report the best linear probing accuracy. This evaluation-time search is applied equally to all variants. Semantic segmentation follows the fixed-layer dense probing protocol specified in the main table without searching.

For CIFAR linear probing, frozen features are globally averaged and passed to a single linear classifier. The probe is trained with horizontal flipping only. For ImageNet linear probing, we use the same frozen-backbone protocol with a parameter-free BatchNorm and a linear classifier, following prior practices. For semantic segmentation, the backbone is frozen and the head contains a LayerNorm and a linear classifier trained without data augmentation. Table~\ref{tab:probe_details} gives more details for the probing heads. These settings are shared by the baseline and feedback variants.

% ==================== Table 10: CIFAR epochs ====================
\begin{table}[t]
    \centering
    \caption{Checkpoint-level Generative Performance on CIFAR}
    \label{tab:cifar_epochs}
    \resizebox{\linewidth}{!}{
    \begin{tabular}{lcccccc}
        \toprule
        \textbf{CIFAR-10} & \multicolumn{2}{c}{\textbf{DDPM, UNet\,(Ho)}} & \multicolumn{2}{c}{\textbf{EDM, UNet\,(Ho)}} & \multicolumn{2}{c}{\textbf{FM, UNet\,(Ho)}} \\
        \cmidrule(lr){2-3}
        \cmidrule(lr){4-5}
        \cmidrule(lr){6-7}
        \textbf{Epoch} & baseline & $m=7$ & baseline & $m=7$ & baseline & $m=7$ \\
        \midrule
400  &        5.31  &           4.97  &        6.01  &           5.67  &        5.42  &           5.28  \\
800  &        4.13  &           3.82  &        4.42  &           4.01  &        4.41  &           4.16  \\
1200 &        3.71  &\underline{3.53} &        3.85  &           3.61  &        4.09  &\underline{3.95} \\
1400 &        3.69  &           3.55  &        3.72  &\underline{3.46} &        4.04  &           3.88  \\
1600 &        3.59  &           3.57  &\textbf{3.54} &           3.37  &        4.00  &           3.83  \\
1800 &\textbf{3.52} &   \textbf{3.45} &        3.55  &   \textbf{3.25} &        3.97  &           3.80  \\
2000 &        3.52  &           3.59  &        3.56  &           3.25  &\textbf{3.94} &   \textbf{3.74} \\
        \bottomrule
    \end{tabular}
    }

\vspace{1ex}

    \resizebox{\linewidth}{!}{
    \begin{tabular}{lcccccc}
        \toprule
        \textbf{CIFAR-10} & \multicolumn{2}{c}{\textbf{DDPM, UNet\,(Song)}} & \multicolumn{2}{c}{\textbf{EDM, UNet\,(Song)}} & \multicolumn{2}{c}{\textbf{FM, UNet\,(Song)}} \\
        \cmidrule(lr){2-3}
        \cmidrule(lr){4-5}
        \cmidrule(lr){6-7}
        \textbf{Epoch} & baseline & $m=9$ & baseline & $m=8$ & baseline & $m=9$ \\
        \midrule
400  &        3.75  &           3.52  &        2.87  &           2.93  &        3.15  &           3.09  \\
800  &        3.16  &           2.99  &        2.38  &           2.27  &        2.68  &\underline{2.59} \\
1200 &        3.03  &\underline{2.83} &\textbf{2.23} &\underline{2.20} &\textbf{2.62} &           2.54  \\
1400 &        3.04  &           2.81  &        2.24  &   \textbf{2.12} &        2.65  &   \textbf{2.52} \\
1600 &\textbf{2.92} &   \textbf{2.79} &        2.26  &           2.15  &        2.65  &           2.57  \\
        \bottomrule
    \end{tabular}
    }

\vspace{2ex}

    \resizebox{\linewidth}{!}{
    \begin{tabular}{lcccccc}
        \toprule
        \textbf{CIFAR-100} & \multicolumn{2}{c}{\textbf{DDPM, UNet\,(Ho)}} & \multicolumn{2}{c}{\textbf{EDM, UNet\,(Ho)}} & \multicolumn{2}{c}{\textbf{FM, UNet\,(Ho)}} \\
        \cmidrule(lr){2-3}
        \cmidrule(lr){4-5}
        \cmidrule(lr){6-7}
        \textbf{Epoch} & baseline & $m=7$ & baseline & $m=6$ & baseline & $m=6$ \\
        \midrule
400  &        9.67  &           9.53  &       11.88  &          11.92  &        9.72  &          10.23  \\
800  &        6.99  &           6.83  &        8.25  &           8.19  &        7.47  &           7.40  \\
1200 &        6.16  &           6.17  &        7.00  &           6.92  &        6.85  &\underline{6.47} \\
1400 &        5.95  &           5.96  &        6.39  &           6.44  &        6.62  &           6.32  \\
1600 &        5.84  &           5.82  &        6.27  &           6.13  &        6.62  &           6.13  \\
1800 &        5.85  &\underline{5.68} &        6.12  &\underline{5.96} &        6.52  &           6.08  \\
2000 &\textbf{5.72} &   \textbf{5.67} &\textbf{6.01} &   \textbf{5.82} &\textbf{6.48} &   \textbf{5.96} \\
        \bottomrule
    \end{tabular}
    }

\vspace{1ex}

    \resizebox{\linewidth}{!}{
    \begin{tabular}{lcccccc}
        \toprule
        \textbf{CIFAR-100} & \multicolumn{2}{c}{\textbf{DDPM, UNet\,(Song)}} & \multicolumn{2}{c}{\textbf{EDM, UNet\,(Song)}} & \multicolumn{2}{c}{\textbf{FM, UNet\,(Song)}} \\
        \cmidrule(lr){2-3}
        \cmidrule(lr){4-5}
        \cmidrule(lr){6-7}
        \textbf{Epoch} & baseline & $m=8$ & baseline & $m=8$ & baseline & $m=9$ \\
        \midrule
400  &        6.47  &           6.46  &        5.90  &           5.97  &        6.11  &           5.90  \\
800  &        4.96  &           4.72  &        4.22  &           4.18  &        4.55  &           4.38  \\
1200 &        4.55  &\underline{4.27} &        3.80  &           3.69  &        4.18  &\underline{4.07} \\
1400 &        4.49  &           4.16  &        3.63  &           3.66  &\textbf{4.16} &           3.97  \\
1600 &\textbf{4.40} &           4.14  &\textbf{3.48} &           3.58  &        4.17  &           4.01  \\
1800 &        4.47  &   \textbf{4.06} &        3.48  &\underline{\textbf{3.37}}&4.16  &   \textbf{3.90} \\

        \bottomrule
    \end{tabular}
    }

\vspace{1.5ex}
\begin{minipage}{0.95\linewidth}
\textbf{Bold} values mark the best FIDs reported in Tab.~\ref{tab:baseline_single}.
\underline{Underlined} values denote the first checkpoint where our model becomes comparable to or better than the baseline best FID, highlighting convergence behavior.
Later UNet\,(Song) results are omitted because FID deteriorates for both models.
\end{minipage}
\end{table}
% ==================== Table 10: CIFAR epochs ====================

\section{Detailed FID Evaluation Results}
\noindent
\textbf{Checkpoint-level FID} on CIFAR and the exact selected depth are reported in Table~\ref{tab:cifar_epochs}. These results complement the best-checkpoint comparison by showing when our model reaches the baseline's best FID under the matched protocol. Our approach demonstrates faster convergence in most cases, and the improvements are not limited to a single checkpoint.
\textbf{Three-seed} results are reported in Table~\ref{tab:cifar_seeds}. Our method consistently reduces FID across three formulations and three runs. For EDM and FM, the mean reductions match the main results. For DDPM, the mean reduction is 0.17, showing a larger, non-negligible gain than suggested by the 0.05 in Tab.~\ref{tab:baseline_single}.

% ==================== Table 11: CIFAR seeds ====================
\begin{table}[t]
    \centering
    \caption{Multi-seed Results on CIFAR-100}
    \label{tab:cifar_seeds}
\resizebox{\linewidth}{!}{
    \begin{tabular}{@{}lcccccc@{}}
        \toprule
        \textbf{CIFAR-100} & \multicolumn{2}{c}{\textbf{DDPM, UNet\,(Ho)}} & \multicolumn{2}{c}{\textbf{EDM, UNet\,(Ho)}} & \multicolumn{2}{c}{\textbf{FM, UNet\,(Ho)}} \\
        \cmidrule(lr){2-3}
        \cmidrule(lr){4-5}
        \cmidrule(lr){6-7}
        \textbf{Runs} & baseline & $m=7$ & baseline & $m=6$ & baseline & $m=6$ \\
        \midrule
        Seed 1\,(Tab.~\ref{tab:baseline_single}) &  5.72  &  5.67  &  6.01  &  5.82  &  6.48  &  5.96  \\
        Seed 2            &  5.64  &  5.34  &  6.00  &  5.72  &  6.71  &  6.16  \\
        Seed 3            &  5.79  &  5.63  &  6.19  &  5.99  &  6.47  &  6.11  \\
        \midrule
        Mean\,$\pm$\,std &  5.72$\pm$0.08  &  5.55$\pm$0.18  &  6.07$\pm$0.11  &  5.84$\pm$0.14  &  6.55$\pm$0.14  &  6.08$\pm$0.10  \\
        \textbf{Reduction} & & \textbf{0.17$\pm$0.13} & & \textbf{0.22$\pm$0.05} & & \textbf{0.48$\pm$0.10} \\
        \bottomrule
    \end{tabular}
}
\end{table}
% ==================== Table 11: CIFAR seeds ====================

\end{document}